\documentclass[runningheads]{llncs}

\usepackage{eccv}

\usepackage{eccvabbrv}
\usepackage{graphicx}
\usepackage{booktabs}
\usepackage{amsmath}
\usepackage{amssymb}
\usepackage{braket}
\usepackage{algorithm}
\usepackage{algorithmic}
\usepackage{soul}
\usepackage{enumitem}
\usepackage{braket}
\usepackage{wasysym}
\usepackage{marvosym}
\usepackage{subcaption}

\makeatletter
\@namedef{ver@everyshi.sty}{}
\makeatother
\usepackage{tikz}
\usetikzlibrary{quantikz2}

\DeclareMathOperator*{\argmin}{arg\,min}

\newcommand{\ba}{\mathbf{a}}

\newcommand{\cK}{\mathcal{K}}

\newcommand{\cD}{\mathcal{D}}

\newcommand{\bF}{\mathbf{F}}
\newcommand{\bff}{\mathbf{f}}

\newcommand{\cI}{\mathcal{I}}

\newcommand{\bu}{\mathbf{u}}

\newcommand{\bp}{\mathbf{p}}

\newcommand{\bs}{\mathbf{s}}

\newcommand{\bt}{\mathbf{t}}

\newcommand{\bv}{\mathbf{v}}

\newcommand{\bx}{\mathbf{x}}

\newcommand{\be}{\mathbf{e}}
\newcommand{\bz}{\mathbf{z}}

\newcommand{\cM}{\mathcal{M}}
\newcommand{\cU}{\mathcal{U}}

\newcommand{\true}[1]{#1^\circ}
\newcommand*{\aka}{\emph{aka}\@\xspace}

\usepackage[accsupp]{axessibility} 

\usepackage{hyperref}
\usepackage{orcidlink}

\begin{document}

\title{Robust Fitting on a Gate Quantum Computer} 

\titlerunning{Robust Fitting on a Gate Quantum Computer}

\author{Frances Fengyi Yang\orcidlink{0000-0003-0731-3563} \and
Michele Sasdelli\thanks{Dr. Michele Sasdelli is funded by Centre for Augmented Reasoning (CAR).}\orcidlink{0000-0003-1021-6369} \and
Tat-Jun Chin\orcidlink{0000-0003-2423-9342}}

\authorrunning{F. F.~Yang et al.}

\institute{The University of Adelaide, Adelaide SA 5000, Australia 
\email{\{fengyi.yang,michele.sasdelli,tat-jun.chin\}@adelaide.edu.au}}

\maketitle

\begin{abstract}
    Gate quantum computers generate significant interest due to their potential to solve certain difficult problems such as prime factorization in polynomial time. Computer vision researchers have long been attracted to the power of quantum computers. Robust fitting, which is fundamentally important to many computer vision pipelines, has recently been shown to be amenable to gate quantum computing. The previous proposed solution was to compute Boolean influence as a measure of outlyingness using the Bernstein-Vazirani quantum circuit. However, the method assumed a quantum implementation of an $\ell_\infty$ feasibility test, which has not been demonstrated. In this paper, we take a big stride towards quantum robust fitting: we propose a quantum circuit to solve the $\ell_\infty$ feasibility test in the 1D case, which allows to demonstrate for the first time quantum robust fitting on a real gate quantum computer, the IonQ Aria. We also show how 1D Boolean influences can be accumulated to compute Boolean influences for higher-dimensional non-linear models, which we experimentally validate on real benchmark datasets.
\end{abstract}

\section{Introduction}\label{intro}

Many computer vision pipelines require estimating a geometric model from noisy and outlier-contaminated data $\cD = \{ \bp_i \}^{N}_{i=1}$. Consensus maximization~\cite{fischler1981random} is a popular framework for robust fitting in computer vision, where one attempts to find the model $\bx$ that agrees with as many of the points as possible, \ie,
\begin{align}\label{eq:maxcon}
\begin{aligned}
\smash{\underset{\bx \in \cM}{\text{maximize}} \sum_{i=1}^N \Psi_\epsilon(r_i(\bx)).}
\end{aligned}
\end{align}
Domain $\cM \subseteq \mathbb{R}^d$ defines the allowable parameters $\bx$ of the model, $r_i : \cM \mapsto \mathbb{R}_{\ge 0}$ is the residual of model $\bx$ w.r.t.~the $i$-th ``point'' $\bp_i$, and
\begin{align}
    \smash{\Psi_\epsilon(r) = \begin{cases} 1 & \textrm{if}~r \le \epsilon, \\ 0 & \textrm{otherwise}, \end{cases}}
\end{align}
is an indicator function that checks if the size of a residual is less than the user-supplied inlier threshold $\epsilon$. Each $\bx$ has a corresponding consensus set
\begin{align}
    \smash{\cI(\bx) = \{ i \in \{1,\dots,N \} \mid r_i(\bx) \le \epsilon \},}
\end{align}
\ie, the set of inliers of $\bx$, and~\eqref{eq:maxcon} is equivalent to finding the $\bx^\ast$ with the largest consensus set $\cI(\bx^\ast)$. The equivalent outlier removal interpretation of~\eqref{eq:maxcon} is
\begin{align}
\begin{aligned}
& \underset{\bx \in \mathcal{M}, \bz \in \{ 0,1 \}^{N}}{\text{minimize}}
& & \| \bz \|_1 \\
& \text{subject to}
& & r_i(\bx) \leq \epsilon \;\; \textrm{if} \;\; z_i = 0, \;\;\;\; i = 1,\dots,N,
\end{aligned}
\end{align}
where $\bz = \left[ z_1, z_2, \dots, z_N \right]$ is a binary vector selecting a subset of $\cD$ as outliers, and the goal is to remove the least number of outliers to find a consensus set.

Arguably the most popular class of methods for~\eqref{eq:maxcon} are the random sampling techniques (\ie, RANSAC~\cite{fischler1981random} and its variants), which do not provide any guarantees. Indeed, complexity results indicate that consensus maximization is generally intractable and inapproximable~\cite{chin2018robust}, and one must resort to exhaustive search to find $\bx^\ast$ (see~\cite{zhang2023accelerating} for a recent survey) or convex relaxation~\cite{Yang22pami-certifiablePerception} to attempt to find a bounded approximation. Consistent with the theory~\cite{chin2018robust}, iterative optimization schemes~\cite{le2017exact,cai2018deterministic} also do not provide optimality guarantees.

More broadly, the difficulty of consensus maximization reflects the general hardness of robust fitting under different formulations and settings~\cite{bernholt2005robust,tzoumas2019outlier}. This has motivated researchers to explore alternative approaches (including {machine learning}~\cite{brachmann2017dsac,ranftl2018deep,truong2021unsupervised,wang2021gdr}) and computing paradigms to solve the problem.

\subsection{Quantum solutions for robust fitting}\label{sec:quantumsolutions}

An active direction has been the development of quantum solutions for robust fitting~\cite{chin2020quantum,doan2022hybrid,Farina_2023_CVPR}. Such methods exploit quantum computers to solve robust fitting or appropriate subproblems within a classical optimization framework.

Chin \etal~\cite{chin2020quantum} proposed to use the Bernstein-Vazirani (BV) quantum circuit to compute Boolean influence, which has proven to be a bona fide measure of outlyingness~\cite{suter2020monotone,Tennakoon_2021_CVPR,Zhang_2022_CVPR}. The BV circuit allows influence computation to be parallelized across data points, thereby yielding a provable speedup. However, the solution rests upon the assumption that an $\ell_\infty$ feasibility test has a quantum implementation. While in theory any classically efficient routine has an efficient quantum realization~\cite[Sec.~3.25]{Nielsen_Chuang_2010}, a quantum implementation of the $\ell_\infty$ test was not provided in~\cite{chin2020quantum}, thereby precluding real quantum demonstrations.

Doan \etal~\cite{doan2022hybrid} exploited the ability of quantum annealers (QA) to conduct energy minimization to build a hybrid quantum-classical robust fitting algorithm. The key is to rewrite~\eqref{eq:maxcon} as a hypergraph vertex cover (HVC) problem, which is amenable to quantum annealing. To avoid the exponential growth in the hyperedges, Doan \etal iteratively sample the hyperedges (on a classical machine) and solve the resulting HVC instances (on a D-Wave QA) to yield intermediate solutions with error bounds. Farina \etal~\cite{Farina_2023_CVPR} proposed quantum multi-model fitting, where quantum annealing is used to solve  disjoint set cover to select the optimal combination of hypotheses that fit the input data.

\subsection{Gate quantum computers vs adiabatic quantum computers}

Gate quantum computers (GQC) and adiabatic quantum computers (AQC), \aka QA, are the two major paradigms to realize quantum computers. Briefly, GQC operates by sequential application of quantum gates to manipulate quantum states encoded in a set of qubits to execute quantum algorithms. On the other hand, AQC follows the adiabatic theorem to gradually evolve a system from an easily prepared initial state to the ground state of a problem Hamiltonian, thereby finding the solution to optimization problems.

GQC is more general than AQC in the sense that the former can implement arbitrary algorithms, potentially with significant speed-ups over the classical counterparts. In fact, some of the most prominent quantum algorithms, \eg, Shor's algorithm for prime factorization~\cite[Chap.~5]{Nielsen_Chuang_2010} and Grover's algorithm for unstructured search~\cite[Chap.~6]{Nielsen_Chuang_2010}, are based on GQC. On the other hand, AQC solves optimization problems in the form of quadratic unconstrained binary optimization (QUBO), hence, the targeted problem needs to be mapped to QUBO form, often by introducing additional variables and hyparameters. The speed-up achievable by AQC is also more difficult to be ascertained~\cite{Villanueva2023why}.

Since it is still unclear which quantum technology will reach maturity (\eg, realizing a million-qubit system that is robust against noise), it is vital for computer vision researchers to explore both pathways. Indeed, both GQC~\cite{chin2020quantum} and AQC~\cite{doan2022hybrid,Farina_2023_CVPR} have been investigated for robust fitting. However, demonstration of~\cite{chin2020quantum} on a GQC is still lacking due to the fundamental gap alluded to above.

\subsection{Contributions}\label{sec:contributions}

This paper bridges the critical gap in~\cite{chin2020quantum} by presenting a novel quantum sub-circuit to conduct the $\ell_\infty$ feasibility test in the BV circuit. Although our solution is limited to 1D problems, it is sufficient to achieve important demonstration of quantum robust fitting on a real GQC, specifically the IonQ Aria~\cite{ionq}.

Further, we show how 1D Boolean influence computation can be embedded in a random sampling framework and accumulated to compute higher-dimensional Boolean influence. This enabled us to demonstrate the validity of the computed influences on real benchmark datasets for two-view geometry estimation.

\subsection{Shortcomings and outlook}\label{sec:shortcomings}

Current gate quantum computers are in the noisy intermediate scale quantum (NISQ) era~\cite{bharti2022noisy}, meaning that the quantum processing units (QPU) contain small number of qubits (tens to low hundreds) and are not sufficiently fault tolerant. This limits the size of input problems and quality of outputs, hence, quantum robust fitting solutions cannot yet outperform established classical methods.

The value of our work lies in exploring an alternative technique that 
\begin{itemize}[topsep=0em]
    \item is theoretically interesting and can inspire novel classical methods, \eg,~\cite{Tennakoon_2021_CVPR,Zhang_2022_CVPR} (note that \emph{quantum computing} is a subject area at ECCV 2024).
    \item could become practical in the medium term, given significant investments into building quantum computers, \eg, IBM's Quantum Roadmap aims to deliver a fully error-corrected system with 200 qubits by 2029~\cite{ibm}.
\end{itemize}

\section{Related work}\label{sec:related}

\subsection{Quantum computing in computer vision}

Quantum computing has been investigated for diverse applications and problems in computer vision (CV). Larasati \etal~\cite{larasati2022trends} provide a comprehensive survey on integrating quantum computing techniques within CV, with a particular emphasis on AQC-assisted algorithms. Their survey highlights key applications in robust fitting \cite{doan2022hybrid}, transformation estimation \cite{meli2022iterative, golyanik2020quantum}, multiple object tracking (MOT) \cite{zaech2022adiabatic}, defect detection in semiconductors \cite{yang2022semiconductor}, and permutation synchronization \cite{birdal2021quantum}. Additional contributions to this field include advancements in motion segmentation \cite{arrigoni2022quantum}, recognition \cite{o2018nonnegative, neven2012qboost}, image classification \cite{cavallaro2020approaching, nguyen2018image, boyda2017deploying, nguyen2020regression}, object detection \cite{li2020quantum}, multi-model fitting \cite{Farina_2023_CVPR}, matching problems \cite{bhatia2023ccuantumm, yurtsever2022q, benkner2020adiabatic, benkner2021q, birdal2021quantum}, and mesh alignment \cite{benkner2021q}. All works mentioned above focus on employing AQC by translating problems into an AQC-admissible form, predominantly QUBO.

Furthermore, there is growing interest in exploring GQC for CV tasks. Research in this area, such as robust fitting~\cite{chin2020quantum} and point set alignment~\cite{noormandipour2022matching}, has laid down a theoretical groundwork, suggesting that quantum-classical hybrid solutions may offer viable paths forward. However, the realization of these concepts in practical quantum implementations has not yet been achieved, with most explorations remaining theoretical or confined to simulations.

\subsection{Quantum computing in machine learning}

The application of quantum computing to assist learning-based approaches for CV tasks has also received significant attention. This approach represents a shift towards integrating the computational capabilities of quantum computing with conventional learning-based methods in CV. Recent examples of such endeavors are as follows: Rosenhahn and Hirche~\cite{rosenhahn2024quantum} tackled anomaly detection by proposing to optimise an ordered set of quantum gates to compute a normalizing flow using quantum architecture search (QAS)~\cite{zhu2022brief}. This work was inspired by a previous study~\cite{liu2018quantum} that formulated a proximity measure to quantify how anomalous a quantum state is. Luo \etal~\cite{luo2024quack} implemented a 10-qubit quantum circuit for binary classification employing stochastic gradient descent within Quantum Machine Learning (QML) and Quantum-Assisted Cluster Kernels (QuACK). Furthermore, Silver \etal~\cite{silver2023mosaiq} introduced MosaiQ, an enhanced generative adversarial network (GAN) framework, building upon~\cite{huang2021experimental}, specifically tailored for the generation of high-quality quantum images. The paper details the quantum circuit ansatz for MosaiQ's generators, which is feasible for execution on contemporary NISQ computers. Training of MosaiQ occurs on a quantum simulator, with inference conducted on both simulator and real QC. \cite{zhang2023quantum} proposed an end-to-end quantum-inspired spectral-spatial pyramid network (QSSPN) for hyperspectral image feature extraction and classification.

These initiatives underscore the promising synergy between quantum computing and learning-based strategies, setting the stage for potential advancements in critical optimization challenges inherent to CV. 

\section{Boolean influence for robust fitting}
\label{sec:bool_robust_fit}

We first review the concept of Boolean influence for robust fitting~\cite{suter2020monotone, Tennakoon_2021_CVPR, Zhang_2022_CVPR}. Recall the binary vector $\bz \in \{0,1 \}^{N}$ that represents selection of a subset of $\cD$, \ie, $z_i = 1$ implies that $\bp_i$ is in subset $\bz$. The \emph{minimax value} of $\bz$ is
\begin{align}\label{eq:minmax}
g(\bz) := \underset{\bx \in \cM}{\text{min}} \; \max_{i=1,\dots,N} \; z_i r_i(\bx) = \underset{\bx \in \cM}{\text{min}}~\left\| \left[ \begin{matrix} z_1 r_1(\bx) \\ \vdots \\ z_N r_N(\bx) \end{matrix} \right] \right\|_\infty,
\end{align}
\ie, $g(\bz)$ is the minimum over $\cM$ of the maximum residual of points in subset $\bz$. We define the $\ell_\infty$ \emph{feasibility test} on $\bz$ as
\begin{align}\label{eq:feasibility}
    f(\bz) := \begin{cases} 0 & \text{if}~g(\bz) \le \epsilon, \\ 1 & \text{otherwise}. \end{cases}
\end{align}

We say that $\bz$ is feasible if $f(\bz) = 0$ and infeasible otherwise. Intuitively, $f(\bz) = 0$ means that there is an $\bx$ to which all the points in $\bz$ have residuals $\le \epsilon$, which in the context of robust fitting implies that $\bz$ is a consensus set. Conversely, $f(\bz) = 1$ implies that $\bz$ contains outliers.

Including new points to a subset cannot make the minimax value of the resultant subset smaller, \ie,
\begin{align}
g(\bz_1) \le g(\bz_1 \vee \bz_2)
\end{align}
where $\vee$ is bit-wise OR. Therefore, $f$ is \emph{monotonic}, \ie,
\begin{align}
f(\bz_1) \le f(\bz_1 \vee \bz_2).
\end{align}
Employing concepts from monotone Boolean function analysis~\cite{odonnell2014}, the Boolean influence (henceforth, just ``influence'') of the $i$-th point under $f$ is
\begin{align}\label{eq:influence}
    \alpha_{i} &= Pr\left[ f(\bz) \ne f(\bz \otimes \be_i ) \right] 
    = \frac{1}{2^N} \sum_{\bz \in \{0,1 \}^N} \mathbb{I}\left[f(\bz) \ne f(\bz \otimes \be_i)\right],
\end{align}
where $\be_i$ is the binary vector of all zeros except at the $i$-th element, $\otimes$ is bit-wise XOR (\ie, $\bz \otimes \be_i$ flips the $i$-th element in $\bz$), and $\mathbb{I}(\cdot)$ returns $1$ if the input condition is true and $0$ otherwise. Intuitively, $\alpha_i$ is the probability that $\bp_i$ will change the feasibility of an arbitrary $\bz$, if $\bp_i$ is inserted or removed from $\bz$.
 
Tennakoon \etal~\cite{Tennakoon_2021_CVPR} and Zhang \etal~\cite{Zhang_2022_CVPR} proved that under certain conditions the influences of the inliers in $\cD$ are strictly smaller than that of the outliers, which supports using influence as a measure of \emph{outlyingness}.

\begin{figure}[t]\centering
     \begin{subfigure}[b]{0.45\columnwidth}\centering
         \includegraphics[width=0.99\columnwidth]{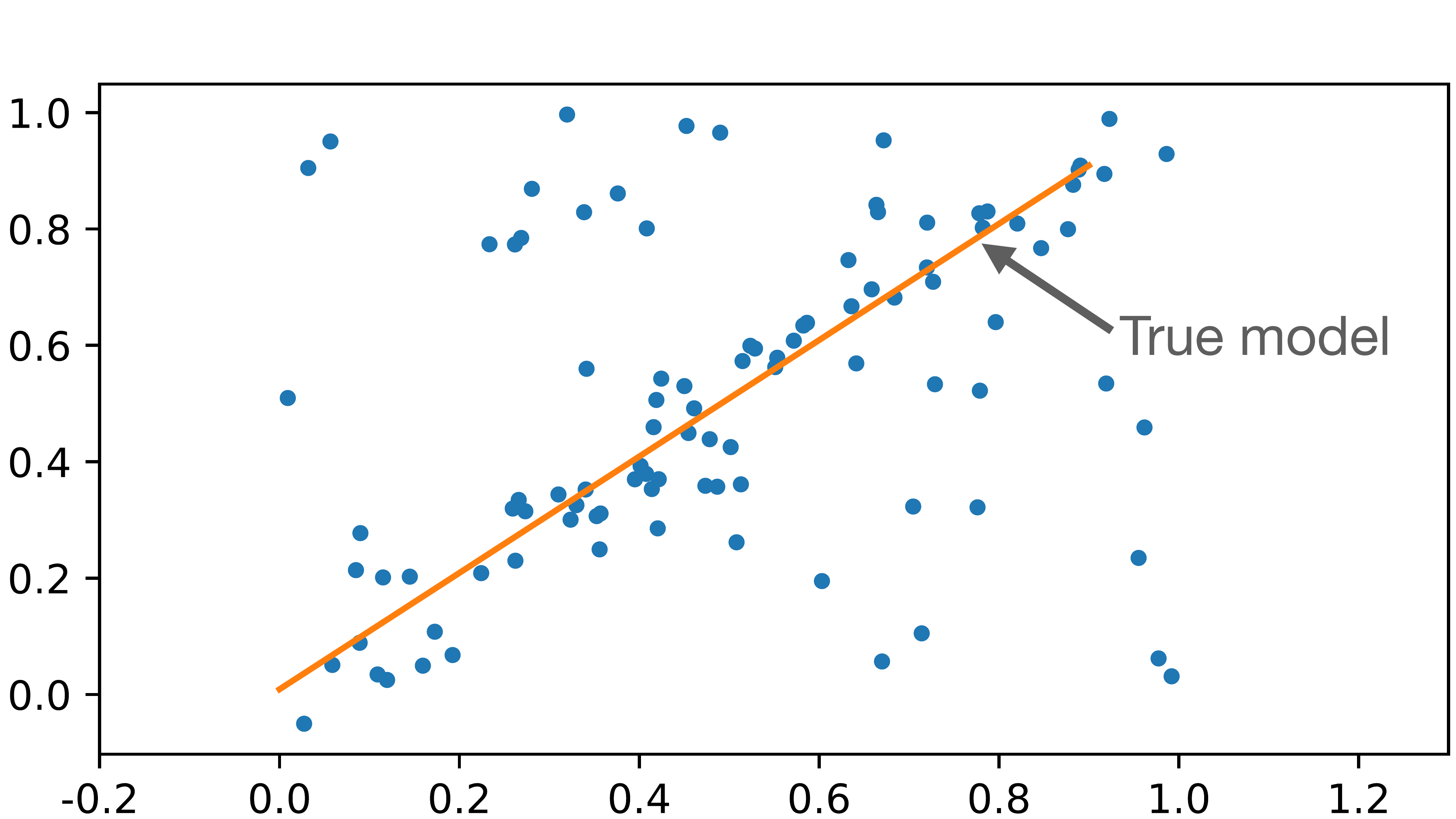}
         \caption{Outlier-prone data $\{ (\ba_i,b_i)\}^{N}_{i=1}$ for linear regression with dimensionality $d = 1$.}
         \label{fig:linreg_data}
     \end{subfigure}
     \begin{subfigure}[b]{0.54\columnwidth}\centering
         \includegraphics[width=0.99\columnwidth]{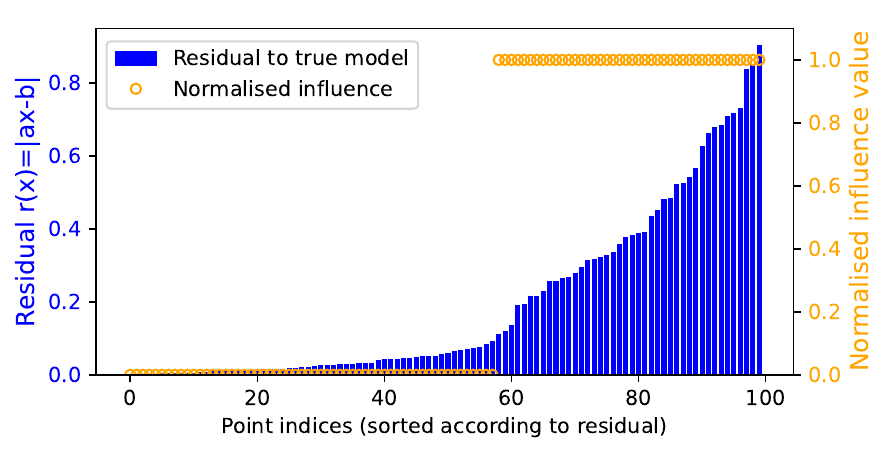}
         \caption{Influences versus residuals to the true model.}
         \label{fig:linreg_infl}
     \end{subfigure}
        \caption{Influences versus true residuals for robust linear regression. Note that points with lower influence tend to have lower true residuals.}
        \label{fig:linreg}
\end{figure}

\begin{example}[Robust linear regression]\label{eg:linreg}
Given $N$ pairs of independent and response measurements $\{ (\ba_i, b_i) \}^{N}_{i=1}$ where $\ba_i \in \mathbb{R}^d$ and $b_i \in \mathbb{R}$, we wish to estimate the linear relationship $\ba^T \bx \approx b$ that best fits the data, which is contaminated with outliers. Define the residual of the $i$-th point $\bp_i = (\ba_i, b_i)$ as
\begin{align}\label{eq:linearres}
    r_i(\bx) = |\ba_i^T \bx - b_i|,
\end{align}
which implies that the minimax problem~\eqref{eq:minmax} reduces to linear programming~\cite{yu14adversarial}. Fig.~\ref{fig:linreg} plots the normalized influences for an outlier-contaminated data for linear regression versus the residuals $r_i(\true{\bx})$ of the data to the true model $\true{\bx}$. Clearly points with lower influences tend to have lower true residuals.
\end{example}

\begin{figure}[htbp]
    \centering
    \begin{subfigure}[t]{0.48\textwidth}
        \centering
        \includegraphics[width=\textwidth]{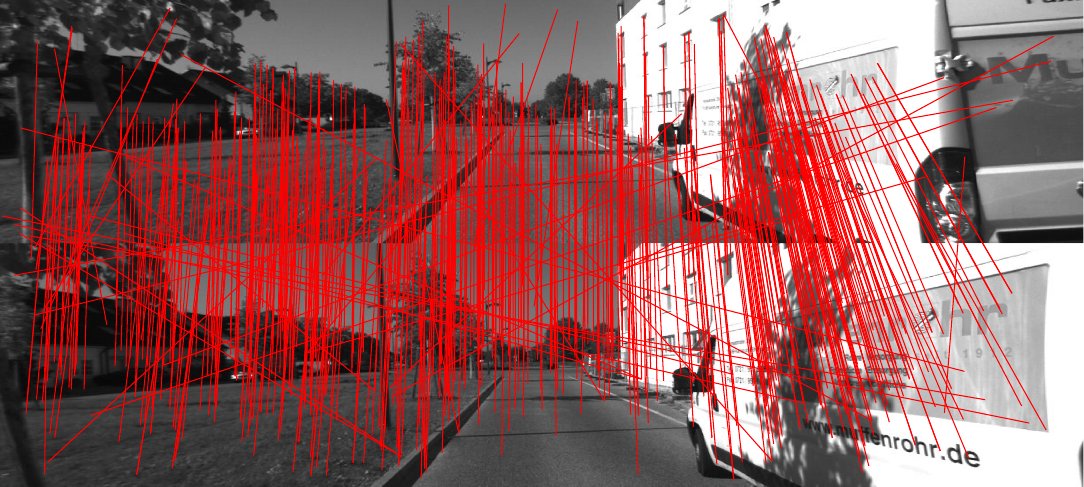}
        \caption{SIFT correspondences~\cite{lowe2004distinctive} on an image pair from KITTI dataset~\cite{bian2019evaluation}.}
    \end{subfigure}%
    \hfill
    \begin{subfigure}[t]{0.51\textwidth}
        \centering
        \includegraphics[width=\textwidth]{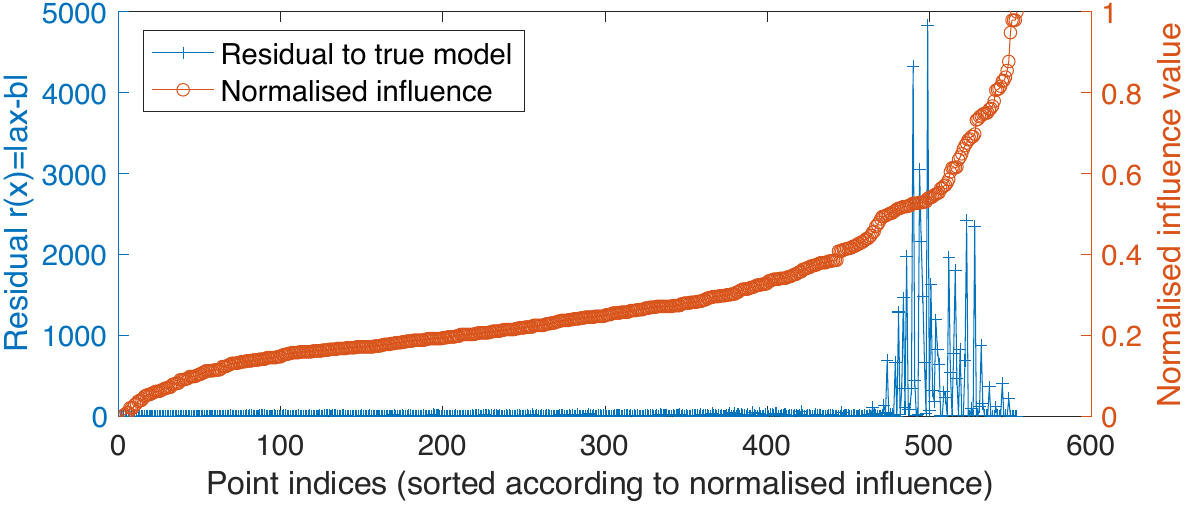}
        \caption{The initial 470 data points, with near-zero residuals, have significantly less influence than the later points with larger residuals.}
    \end{subfigure}%
    \caption{Influences versus true residuals for a correspondence set for fundamental matrix estimation. Note that points with lower influence tend to have lower true residuals.}
    \vspace{-5pt}
    \label{fig:fundamental}
\end{figure}

\begin{example}[Fundamental matrix estimation]\label{eg:fundmatrix}
We aim to find the fund.~matrix $\bF$, a homogeneous $3\times 3$ matrix of rank $2$, that defines the epipolar equation
\begin{align}\label{eq:epipolar}
    (\tilde{\bu}^\prime)^T\bF \tilde{\bu} = 0
\end{align}
between two views, where $\langle \bu, \bu^\prime \rangle$ is a point correspondence with $\bu = \left[ \begin{matrix} u & v \end{matrix} \right]^T$, $\bu^\prime = \left[ \begin{matrix} u^\prime & v^\prime \end{matrix} \right]^T$, and $\tilde{\bu} = \left[ \begin{matrix} \bu^T & 1 \end{matrix} \right]^T$. Based on~\cite[(11.2)]{Hartley2004multiple}, we linearize~\eqref{eq:epipolar} to
\begin{align}\label{eq:linearise}
\left[u^\prime u, u^\prime v, u^\prime, v^\prime u, v^\prime v, v^\prime, u, v, 1\right] \bff = 0,
\end{align}
where $\bff \in \mathbb{R}^9$ is vectorized $\bF$. Following~\cite[Sec.~4.1.2]{Hartley2004multiple}, we dehomogenize~\eqref{eq:linearise} by fixing the first element of $\bff$ to $1$, and moving $u^\prime u$ to the RHS to yield
\begin{align}\label{eq:dehomogenise}
\left[u^\prime v, u^\prime, v^\prime u, v^\prime v, v^\prime, u, v, 1\right] \bx = -u^\prime u \;\; \Longrightarrow \;\; \ba \bx = b,
\end{align}
where $\bx \in \mathbb{R}^8$ contains the rest of $\bff$. Given a set of outlier-prone correspondences $\{ \langle \bu_i, \bu_i^\prime \rangle \}^{N}_{i=1}$, we take $\bp_i = (\ba_i,b_i)$ following~\eqref{eq:dehomogenise}, and adopt~\eqref{eq:linearres} as the residual. In short, we have converted fundamental matrix estimation into linear regression. Fig.~\ref{fig:fundamental} plots the normalized influences for an outlier-contaminated set of correspondences versus the residuals $r_i(\true{\bx})$ of the correspondences to the true model $\true{\bx}$, which was derived from the true fundamental matrix $\true{\bF}$. Evidently points with lower influences have lower true residuals.
\end{example}

\subsubsection{Model estimation}
\label{subsub:model_est1}

Given the influences $\{ \alpha_i \}^{N}_{i=1}$, they can be thresholded to yield an inlier set, on which least squares can be applied to estimate  $\cM$~\cite{chin2020quantum}. More sophisticated heuristics guided by the influences have also been proposed~\cite{Tennakoon_2021_CVPR,Zhang_2022_CVPR}.

The viability of influence as an outlying measure on other non-linear estimation problems (\eg, homography estimation, 3D triangulation) has also been established~\cite{chin2020quantum,Tennakoon_2021_CVPR,Zhang_2022_CVPR}. However, the bottleneck lies in computing the influence~\eqref{eq:influence}.

\section{Quantum algorithm for influence computation}

Note that~\eqref{eq:influence} sums over all $2^N$ combinations of $\bz$. In practice, a finite sample set $\mathsf{Z} = \{ \bz_j \}^{M}_{j=1} \subset \{0,1\}^N$ of size $M$ is procured to approximate the influence
\begin{align}
    \smash{\hat{\alpha}_i = \frac{1}{M}\sum^{M}_{j=1} \mathbb{I}\left[f(\bz_j) \ne f(\bz_j \otimes \be_i)\right].}
\end{align}
Alg.~\ref{alg:classical_influence} summarizes the method for influence approximation. Note that Figs.~\ref{fig:linreg} and~\ref{fig:fundamental} plot approximate influences for $M = 1000$. It can be shown~\cite{chin2020quantum} that $\hat{\alpha}_i$ approaches $\alpha_i$ following the probabilistic bound
\begin{align}
    \smash{Pr(|\hat{\alpha}_i - \alpha_i | < \delta) > 1 - 2e^{-2M\delta^2},}
\end{align}
where $\delta$ is the desired deviation. For example, if $M = 1000$, $Pr(|\hat{\alpha}_i - \alpha_i | < 0.05) > 0.99$, \ie, not many samples are required to achieve a good approximation. However, Alg.~\ref{alg:classical_influence} can be costly if $N$ is large, \eg, $N > 1000$ points.

\subsection{Quantum algorithm}
\label{sec:quantum_alg}

\begin{figure}[ht]\centering
\vspace{-10pt}
\includegraphics[width=.5\columnwidth]{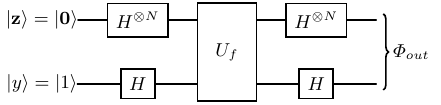}
\vspace{-5pt}
\caption{BV quantum circuit with generic $U_f$ for influence computation.}\label{fig:bv}
\vspace{-10pt}
\end{figure}

Chin \etal~\cite{chin2020quantum} proposed a quantum algorithm to speed up (approximate) influence calculation; Alg.~\ref{alg:quantum_influence} summarizes the method. The algorithm employs the BV circuit~\cite{bernstein1993quantum} (shown in Fig.~\ref{fig:bv}), which we briefly describe below; see~\cite{chin2020quantum} for details.

\begin{algorithm}[t]
\begin{algorithmic}[1]
    \REQUIRE $N$ input data points $\cD$, inlier threshold $\epsilon$, number of iterations $M$.
    \FOR{$j = 1,\dots,M$}
    \STATE $\bz_j \leftarrow$ Randomly sample a subset of $\{1,\dots,N\}$.
    \FOR{$i = 1,\dots,N$}
    \STATE $s_{j,i} \leftarrow \mathbb{I}\left[f(\bz_j) \ne f(\bz_j \otimes \be_i)\right]$.
    \ENDFOR
    \ENDFOR
    \FOR{$i = 1,\dots,N$}
    \STATE $\hat{\alpha}_i \leftarrow \frac{1}{M} \sum_{j=1}^{M} s_{j,i}$.
    \ENDFOR
    \RETURN $\{ \hat{\alpha}_i \}^{N}_{i=1}$.
\end{algorithmic}
\caption{Classical method for influence approximation.}\label{alg:classical_influence}
\end{algorithm}

\begin{algorithm}[t]
\begin{algorithmic}[1]
    \REQUIRE $N$ input data points $\cD$, inlier threshold $\epsilon$, number of iterations $M$.
    \FOR{$j = 1,\dots,M$}
    \STATE $\left[ s_{j,1}, s_{j,2}, \dots, s_{j,N} \right] \leftarrow$ Run BV circuit with $\cD$ and $\epsilon$ and measure top-$N$ qubits.
    \ENDFOR
    \FOR{$i = 1,\dots,N$}
    \STATE $\hat{\alpha}_i \leftarrow \frac{1}{M} \sum_{j=1}^{M} s_{j,i}$.
    \ENDFOR
    \RETURN $\{ \hat{\alpha}_i \}^{N}_{i=1}$.
\end{algorithmic}
\caption{Quantum method for influence approximation.}\label{alg:quantum_influence}
\end{algorithm}

The adopted BV circuit requires $N+1$ qubits: the top-$N$ qubits $\ket{\bz}$ represent 
subset selection and the bottom qubit $\ket{y}$ is for auxiliary purposes. The major 
steps of the BV circuit in the context of Alg.~\ref{alg:quantum_influence} are:
\begin{description}
    \item[Initialization] Set $\ket{\bz} = \ket{\mathbf{0}}$ and $\ket{y} = \ket{1}$, then pass them through Hadamard gates to generate uniform superposition for the top-$N$ qubits
    \begin{align}\label{eq:uniform}
        \ket{\bz} = \frac{1}{\sqrt{2^N}} \sum_{\bt \in \{0,1\}^N} \ket{\bt},
    \end{align}
    while the auxiliary qubit is prepared in a state that will facilitate interference.
    \item[Oracle function] The quantum oracle, represented by gate $U_f$, implements the feasibility test~\eqref{eq:feasibility} corresponding to the target problem (more in Sec.~\ref{sec:approach}).
    \item[Interference] After the oracle function, another set of Hadamard gates perform constructive and destructive interference, resulting in the output state
    \begin{align}
        \Phi_{out} = \frac{1}{2^N} \sum_{\bs \in \{0,1\}^N} \sum_{\bt \in \{0,1 \}^N} (-1)^{f(\bt)+\bs\cdot\bt}\ket{\bs}\ket{1}.
    \end{align}
    \item[Measurement] Measure the top-$N$ qubits of $\Phi_{out}$ to yield binary string $\bs$.
\end{description}
As outlined in~\cite[Theorem 1]{chin2020quantum}, each element $s_i$ in the measured binary string $\bs = [s_1,s_2,\dots,s_N] \in \{0,1\}^N$ is a sample from the Bernoulli distribution
\begin{align}
Pr(s_i = 1) = \alpha_i.
\end{align}
By running the BV circuit $M$ times, Alg.~\ref{alg:quantum_influence} effectively acquires outcomes of the test $\mathbb{I}\left[f(\bz_j) \ne f(\bz_j \otimes \be_i)\right]$ for $M$ samples $\{ \bz_j \}^{M}_{j=1}$. Moreover, the sampling is parallelized across the $N$ points by the QPU, thus, Alg.~\ref{alg:quantum_influence} can have a significant computational gain over Alg.~\ref{alg:classical_influence} if $N$ is large.

\subsection{Quantum circuit for $\ell_\infty$ feasibility test}\label{sec:approach}

Kahl and Hartley~\cite{kahl2008multiple} showed that the $\ell_\infty$ feasibility test~\eqref{eq:feasibility} is efficient for quasiconvex residuals (which include the linear case in Examples~\ref{eg:linreg} and~\ref{eg:fundmatrix}). Chin \etal~\cite{chin2020quantum} relied on the fact that any classical computation has a comparably efficient quantum implementation~\cite[Chap.~3.25]{Nielsen_Chuang_2010} to argue for the existence of $U_f$, which is responsible for conducting the $\ell_\infty$ feasibility test on the quantum machine; however, no concrete solution was provided for $U_f$, which raises doubts of the feasibility of Alg.~\ref{alg:quantum_influence} on a real gate quantum computer.

In this section, we address the critical gap by proposing a quantum circuit for $U_f$. Although our solution is aimed at the simple 1D ``point fitting'' problem,
\begin{itemize}[topsep=0em]
    \item it is a non-trivial and novel contribution that enables quantum robust fitting to be demonstrated for the first time on a real gate quantum computer.
    \item the computed influences for point fitting can be accumulated to achieve influence computation for more complex estimation tasks (Sec.~\ref{sec:multiD}).
\end{itemize}

\subsubsection{Robust point fitting}

As alluded to above, our proposed quantum circuit is aimed at robust point fitting, which is a 1D special case of robust linear regression (Example~\ref{eg:linreg}) with $d = 1$ and $a_i = 1, \forall i$. The residual reduces to
\begin{align}
    \smash{r_i(x) = |x - b_i|,}
\end{align}
and we aim to find the point $x \in \mathbb{R}$ that is representative (within $\epsilon$) of the inlier population in the data $\cD = \{ b_i \}^{N}_{i=1}$. The $\ell_\infty$ feasibility test~\eqref{eq:feasibility} reduces to
\begin{align} \label{eq:feasibility_simplified}
    \smash{f(\bz) = \begin{cases} 0 & \text{if}~\max( \{ b_i \mid i \; \text{s.t.} \; z_i = 1\}) - \min(\{ b_i \mid i \; \text{s.t.} \; z_i = 1 \}) \le 2\epsilon, \\ 1 & \text{otherwise}. \end{cases}}
\end{align}

\subsubsection{Preprocessing}
We first sort $\cD$ decreasingly such that $b_i \ge b_j$ $\forall$ $i < j$. Also, we assume $b_i \ge 0$ $\forall$ $i$, by offsetting the values as necessary such that $b_N = 0$.

\subsubsection{Circuit design}
\begin{figure}[t]\centering
    \vspace{-1em}
    \includegraphics[width=0.7\textwidth]{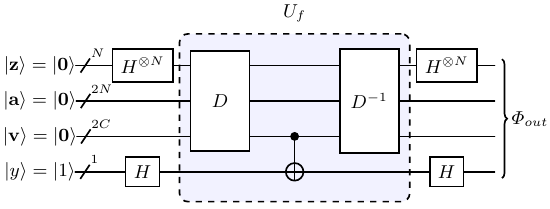}
    \caption{Proposed quantum gate $U_f$ for point fitting embedded in the BV circuit. The $U_f$ gate introduces auxiliary qubits $\ket{\ba}$ and $\ket{\bv}$. The subcircuit $D$ is illustrated in Fig.~\ref{fig:detailed}.}
    \label{fig:overview}
\end{figure}

\begin{figure}[t]\centering
    \includegraphics[width=.7\columnwidth]{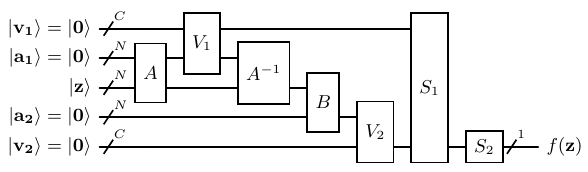}
    \caption{Composition of subcircuit $D$ from Fig.~\ref{fig:overview}, which includes largest number selector $A$, smallest number selector $B$, subcircuits $V_1$ and $V_2$ to encode the data values $\{ b_i \}^{N}_{i=1}$, a quantum Fourier transform (QFT) subtractor $S_1$~\cite{ruiz2017quantum,draper2000addition}, and finally a $2\epsilon$ comparator $S_2$. $f(\bz)$ is indicated by the last bit of $\ket{\bf{v_2}}$. Details of $A$, $B$, $S_1$, and $S_2$ are in the supp.~material, while an example of $V_1$ and $V_2$ is given in Fig.~\ref{fig:v1v2}.}
    \vspace{-15pt}
    \label{fig:detailed}
\end{figure}

\begin{figure}[t]\centering
    \includegraphics[width=.5\columnwidth]{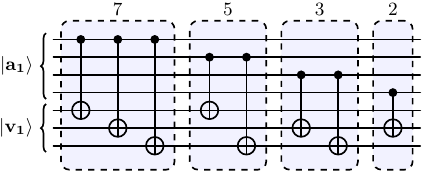}
    \caption{An example of $V_1$ that encodes the data $\cD = \{7,5,3,2\}$ for point fitting. The number of qubits employed for $\ket{\bv_1}$ reflects the precision used for the data. The exact same design applies to $V_2$, save for changing $\ket{\ba_1}$ to $\ket{\ba_2}$ and $\ket{\bv_1}$ to $\ket{\bv_2}$.}
    \label{fig:v1v2}
\end{figure}

The proposed $U_f$ is shown in Fig.~\ref{fig:overview}, with the implementation of subcircuit $D$ shown in Fig.~\ref{fig:detailed}. Our design introduces ancillary qubits $\ket{\ba} = \ket{\ba_1 \ba_2}$ and $\ket{\bv} = \ket{\bv_1 \bv_2}$, which are all initialized to $\ket{\mathbf{0}}$. The major circuit blocks are:
\begin{itemize}[topsep=0em,leftmargin=1em]
    \item $D$ is the main circuit that computes $f(\bz)$, while $D^{-1}$ performs ``uncomputation'' to restore the ancillary qubits as required for $U_f$ in the BV circuit.
    \item $A$ selects the largest (sorted) number in the subset indicated by $\ket{\bz}$. For example, for input $\cD = \{7,5,3,2\}$ and $\ket{\bz} = \ket{1011}$, applying $A$ yields $\ket{\ba_1} = \ket{1000}$, implying the size of $\ket{\ba_1}$ being the number of data points $N$. Block $A^{-1}$ reverses the computation such that $\ket{\bz}$ and $\ket{\ba_1}$ are restored. Note that following the sequence in Fig.~\ref{fig:overview}, $\ket{\bz}$ is in uniform superposition~\eqref{eq:uniform} when input to $A$.
    \item $B$ selects the smallest (sorted) number in the subset indicated by $\ket{\bz}$, \eg, for input $\cD = \{7,5,3,2\}$ and $\ket{\bz} = \ket{1011}$, applying $B$ yields $\ket{\ba_2} = \ket{0001}$.
    \item $V_1$ and $V_2$ encode the numerical values of the input data $\cD$. The number of qubits in $\ket{\bv_1}$ and $\ket{\bv_2}$ equals the bit precision $C$ of the data. An example data entry to $V_1$ and $V_2$ is demonstrated in Fig. \ref{fig:v1v2}.
    \item $S_1$ calculates the absolute difference between the values encoded in $\ket{\bv_1}$ and $\ket{\bv_2}$ and selected by $\ket{\ba_1}$ and $\ket{\ba_2}$. The output is encoded in $\ket{\bv_2}$. The subtractor~\cite{ruiz2017quantum,draper2000addition} is based on {Quantum Fourier transform (QFT)}~\cite{coppersmith2002approximate}.
    \item $S_2$ takes in $\ket{\bv_2}$ and calculates its absolute difference with an inbuilt value of $2\epsilon$, which effectively conducts the feasibility test $f(\bz)$ \eqref{eq:feasibility_simplified}.
\end{itemize}
For brevity, we provide the detailed design of the circuits in the supp.~material.

\subsubsection{Qubit count}

Denoting $N$ as the number of points and $C$ as the bit precision of the data, the circuit in Fig.~\ref{fig:overview} consumes $3N+2C+1$ qubits. Further optimization is possible to reduce the qubit count---see supplementary material.

\section{Influence-based robust fitting of fundamental matrix}\label{sec:multiD}

Here, we show how influences computed for robust point fitting, which our quantum circuit is specialized for (Sec.~\ref{sec:approach}), can be accumulated to obtain influences for more complex robust fitting tasks, specifically fundamental matrix estimation. Alg.~\ref{alg:multiD} summarizes the proposed method. Note that the main goal of Alg.~\ref{alg:multiD} is \emph{enabling validation of the computed and accumulated influences} (Sec.~\ref{sec:fundmatrixresults}).

\begin{algorithm}[t]
\begin{algorithmic}[1]
    \REQUIRE Point correspondences $\cU = \{ \langle \bu_i, \bu^{\prime}_i \rangle \}^{N}_{i=1}$, inlier threshold $\epsilon$, number of hypotheses $T$, number of iterations $M$.
    \item $\cD = \{(\ba_i,b_i)\}^{N}_{i=1} \leftarrow$ Linearize the measurements $\cU$ (see Example~\ref{eg:fundmatrix}).
    \FOR{$t = 1,\dots,T$}
    \STATE $\cK_t \leftarrow$ Sample a minimal subset of 8 correspondences from $\cU$.\label{step:hypsampling1}
    \STATE $\bF_t \leftarrow$ Estimate fundamental matrix from $\cK_t$ using 8-point method~\cite{Hartley2004multiple}.\label{step:hypsampling2}
    \STATE $\{ r_i \}^{N}_{i=1} \leftarrow$ Evaluate linearized residual of $\cD$ on $\bF_t$ (see Example~\ref{eg:fundmatrix}).\label{step:residual}    
    \STATE $\{ \hat{\alpha}_{i}^{(t)} \}^{N}_{i=1} \leftarrow$ Run Alg.~\ref{alg:classical_influence} or~\ref{alg:quantum_influence} for $M$ iterations with threshold $\epsilon$ to compute influences for point fitting problem with residuals $\{ r_i \}^{N}_{i=1}$ as data.\label{step:influence}    
    \ENDFOR
    \FOR{$i = 1,\dots,N$}
    \STATE $\hat{\alpha}_i \leftarrow \frac{1}{T} \sum_{t=1}^{T} \log\left( \hat{\alpha}^{(t)}_{i}\right)$.\label{step:accum}
    \ENDFOR
    \RETURN $\{ \hat{\alpha}_i \}^{N}_{i=1}$.
\end{algorithmic}
\caption{Influence accumulation for fundamental matrix estimation.}\label{alg:multiD}
\end{algorithm}

The main idea is conducting RANSAC-style hypothesis sampling (Steps~\ref{step:hypsampling1}--\ref{step:hypsampling2}), evaluate the residual of the input correspondences on each hypothesis (Step~\ref{step:residual}), then use the residuals as input data for point fitting influence computation (Step~\ref{step:influence}). The point fitting influences are then accumulated via log averaging (Step~\ref{step:accum}) to obtain the influences for robust fundamental matrix fitting.

Due to the limitations of current quantum hardware (Sec.~\ref{sec:shortcomings}), the number of points $N$ that are feasible for the quantum method (Alg.~\ref{alg:quantum_influence}) is restricted (see Sec.~\ref{sec:qpuresults}). Thus, to verify Alg.~\ref{alg:multiD}, we will mainly use the classical method (Alg.~\ref{alg:classical_influence}).

\vspace{-.5em}
\subsubsection{Model estimation}
\label{subsub:model_est2}

We depart from previous influence-based estimation methods~\cite{chin2020quantum,Tennakoon_2021_CVPR,Zhang_2022_CVPR}. We first normalize the influences $\{ \hat{\alpha}_i \}^{N}_{i=1}$ to $[0,1]$. For each $\gamma_h$ from a uniform sample of thresholds $[\gamma_1,\gamma_2,\dots,\gamma_H] \subset [0,1]$, we solve
\begin{align}
\label{eq:model_selection}
    \smash{\bx^{\ast}_h = \argmin_{\bx \in \cM} \sum^{N}_{i=1} \mathbb{I}(\hat{\alpha}_i \le \gamma_h) r_i(\bx)^2,}
\end{align}
\ie, least squares fitting on points with influence $\le \gamma_h$. The model $\bx^{\ast}_h$ 
with the highest consensus $\cI(\bx^{\ast}_h)$ is returned as the final robust estimate.

\section{Experiments}\label{sec:experiments}
\subsection{Correctness and feasibility of quantum algorithm}\label{sec:qpuresults}
\subsubsection{Correctness of quantum circuit}

\begin{figure}[h]\centering
        \vspace{-1.5em}
         \includegraphics[width=0.65\columnwidth]{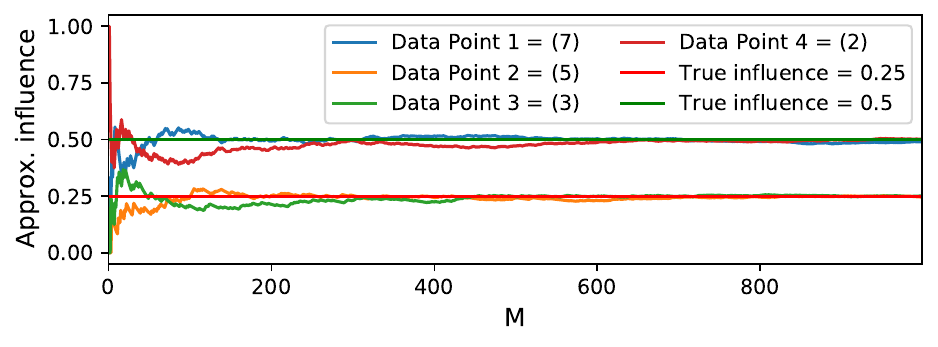}
         \vspace{-1.5em}
        \caption{Approximate influence as a function of iteration count $M$ in Alg.~\ref{alg:quantum_influence}, for data $\cD = \{7,5,3,2\}$ and $\epsilon = 1$ and implemented on Amazon Braket SV1 quantum simulator.}        
        \label{fig:influenceplot}
\end{figure}

To validate the proposed quantum algorithm (Sec.~\ref{sec:approach}), we used the State Vector Simulator (SV1) on Amazon Braket~\cite{sv1}, which contained 34 qubits. Via the Qiskit framework~\cite{Qiskit}, we implemented the proposed quantum circuit (Fig.~\ref{fig:overview}) with input data $\cD = \{7,5,3,2\}$ at $C = 3$ bit precision and $\epsilon = 1$, which consumed 20 qubits (see Sec.~\ref{sec:approach}). Fig.~\ref{fig:influenceplot} plots the computed influences as a function of iteration count $M$ in Alg.~\ref{alg:quantum_influence}. It is evident that the computed influences approached the true values $\{0.50,0.25,0.25,0.50\}$ with increasing $M$. The results indicate the correctness of our circuit and Alg.~\ref{alg:quantum_influence}.

\vspace{-2.5em}
\subsubsection{Feasibility on quantum hardware}

To demonstrate the feasibility of our proposed quantum algorithm (Sec.~\ref{sec:approach}) on a real gate quantum computer, we tested it on IonQ Aria, which is IonQ's 5th generation trapped-ion quantum computer with $25$ physical qubits. The proposed quantum circuit (Fig.~\ref{fig:overview}) with several $\cD$ and $\epsilon$ (see Tab.~\ref{tab:influence}) were executed on Aria and SV1 for comparison. Due to limited access to the Aria, only $M = 50$ iterations were executed (with reported runtimes of seconds), as compared to $M = 1000$ on SV1.

Tab.~\ref{tab:influence} shows that the approximate influences from SV1 aligned closely with the true values. On Aria, the approximate influences were close to the true values when the circuit size was small; however, as the circuit size increased, the discrepancy with the true values also increased. This could be due to:
\begin{itemize}[topsep=0em]
    \item Insufficient number of iterations $M$ for Alg.~\ref{alg:quantum_influence} on the Aria.
    \item Lack of error correction on the quantum hardware, which was difficult to implement anyway due to low QPU capacity. Note that the sizes of the larger circuits tested were close to the maximum number of qubits (25).
\end{itemize}
Nevertheless, the good results from the smaller circuits establish the feasibility of Alg.~\ref{alg:quantum_influence} and the proposed quantum circuit (Fig.~\ref{fig:overview}) on  real quantum hardware.

\begin{table}[!htp]\centering
\caption{Results of Alg.~\ref{alg:quantum_influence} on a real gate quantum computer (IonQ Aria) for $M = 50$ runs and a quantum simulator (Amazon Braket SV1) for $M = 1000$. Legend: $N$ is the number of points, $C$ is the bit precision, $\cD$ is input data, $nQ$ is the number of qubits required to implement the quantum circuit (Fig.~\ref{fig:overview}), $2\epsilon$ is twice the inlier threshold, $\{ \alpha_i \}^{N}_{i=1}$ are the true influences, $\{ \hat{\alpha}^{Q}_i \}^{N}_{i=1}$ are the approx.~influence from executing Alg.~\ref{alg:quantum_influence} on IonQ Aria, $\{ \hat{\alpha}^{S}_i \}^{N}_{i=1}$ are the approx.~influence from executing Alg.~\ref{alg:quantum_influence} on SV1.}\label{tab:influence}
\scriptsize
\begin{tabular}{|r|r|r|r|r|r|r|r|r|}
\hline
$N$ &$C$ &$\cD$ & $nQ$ & $2\epsilon$ & $\{ \alpha_i \}^{N}_{i=1}$ & $\{ \hat{\alpha}^{Q}_i \}^{N}_{i=1}$ & $\{ \hat{\alpha}^{S}_i \}^{N}_{i=1}$ \\
\hline
2 &1 &\{0, 1\} &9 &0 & $0.50, 0.50$  &$0.50, 0.56$  &$0.51, 0.49$\\
2 &1 &\{0, 1\} &9 &1 & $0, 0$  &$0,0$  &$0,0$\\
2 &2 &\{0, 2\} &11 &1 & $0.50, 0.50$  &$0.50,0.58$  &$0.50,0.49$\\
2 &3 &\{2, 4\} &13 &1 & $0.50, 0.50$  &$0.46,0.56$  &$0.51,0.50$\\
2 &3 &\{2, 4\} &13 &2 & $0, 0$  &$0,0$ &$0,0$\\
3 &3 &\{2, 4, 7\} &16 &3 & $0.50 ,0.25, 0.50$  &$0.49,0.23,0.53$  &$0.50,0.26,0.52$\\
4 &3 &\{2, 3, 5, 7\} &19 &2 & $0.50, 0.25, 0.25, 0.50$  &$0.48,0.39,0.41,0.53$  &$0.50,0.25,0.25,0.49$\\
\hline
\end{tabular}
\vspace{-1.5em}
\end{table}

\subsection{Influence-based robust fitting of fundamental matrix}\label{sec:fundmatrixresults}

{We designed an experiment to validate the usefulness of the accumulated influences for a complex robust fitting task. Provided an efficient way to compute influences (e.g. with quantum computing), we showed that 
competitive CV algorithms can be built. }
We evaluated the method described in Sec.~\ref{sec:multiD} on the image pairs collected by \cite{bian2019evaluation} from the  TUM\cite{sturm2012benchmark}, KITTI\cite{geiger2012we}, T\&T\cite{knapitsch2017tanks} and CPC\cite{wilson2014robust} datasets.
Each collection has $1000$ image pairs with views of the same scene and a ground truth fundamental matrix between the views.
We used standard established techniques for keypoint extraction (SIFT\cite{lowe2004distinctive}) and matching proposals\cite{lowe2004distinctive}, returning typically hundreds, up to a few thousands correspondences.
The fundamental matrix estimation problem was linearized following Example~\ref{eg:fundmatrix}.
We constructed ground truth inliers/outliers labels. 
For each correspondence we calculated the epipolar lines and thresholded the sum of the two distances between pixels and the lines at $6$ pixels to create the labels.

First, the influences for the correspondences were obtained by  Alg.~\ref{alg:multiD} together with Alg.~\ref{alg:classical_influence} for influence computation.
As shown by \cite{chin2020quantum}, influences can be used to distinguish inliers from outliers. A Receiver Operating Characteristic (ROC) Area Under the Curve (AUC) score was calculated for each image pair. The averages of the score over the datasets are shown in Tab.~\ref{table:roc}. We report the classification ROC AUC score of a RANSAC estimator and of the influences estimated as in \cite{chin2020quantum}. Interestingly, the new method scored favorably compared to both RANSAC residuals and the original method from \cite{chin2020quantum}.

We then focused on testing the capabilities of using influences for fundamental matrix estimation. To calculate the model we used the method summarized by \eqref{eq:model_selection} in Sec.~\ref{subsub:model_est2}. For the evaluation of the fundamental matrices we use the Symmetric Geometric Distance (SGD)\cite{zhang1998determining} in pixels between virtual correspondences built with the estimated matrix and the ground truth matrix.
We followed\cite{bian2019evaluation} and normalise SGD between $[0,1]$ by the length of the image diagonal $f = 1 / \sqrt{(w^ + h^2)}$ obtaining Normalised SGD (NSGD).
The NSGD is a method to measure the distance between the ground truth fundamental matrix and the estimated fundamental matrix. Values close to $0$ are optimal.  
When the estimated fundamental matrix has a NSGD lower that $0.05$ it is considered accurate.   
We evaluated the models with the recall of correctly estimated models ($recall = \frac{\#correct}{\#total}$). The performances are reported in Tab.~\ref{table:SGD}. On the left half of the table we also reported RANSAC and USAC-openCV estimators run on the same set, with the same number of iterations and the same inlier/outlier threshold used with our method for fair comparison. On the right half of the table we collected the results from\cite{bian2019evaluation} for easy comparison.
Our approach showed that an efficient way of calculating influences can be beneficial for robust fitting. 

\vspace{-.2em}
\subsubsection{Hyperparameters}

In the above fundamental matrix experiments, the hyperparameters for our method (Sec.~\ref{sec:multiD}) were set as follows: The threshold of the linearized problem was set to $\epsilon=0.6$; $M$ was set to $1000$; $T$ was set to $1000$; and $H=50$ was the number of influence thresholds tested.

\vspace{-1em}
\begin{table}\centering
\caption{Average ROC AUC classification scores of inliers/outliers by RANSAC, influence estimated using Alg.~\ref{alg:multiD}, and influence estimated using ~\cite{chin2020quantum} on the linearized problem. The expectation of a random classifier is $0.5$ while a perfect classifier returns $1$; hence, higher ROC AUC indicate better performance.}
\label{table:roc}
\vspace{-5pt}
\begin{tabular}{c|c|c|c}
                 & RANSAC & Influence (Alg.~\ref{alg:multiD}) & Influence\cite{chin2020quantum}    \\  
\hline
TUM              & 0.71 & 0.83  & 0.76   \\ 
KITTI            & 0.86 & 0.94 & 0.86  \\ 
T\&T             & 0.72 & 0.82 & 0.82  \\ 
CPC              & 0.75 & 0.87 & 0.85  \\ 
\end{tabular}
\end{table}

\vspace{-2.5em}
\begin{table}\centering
\caption{Percentage recall at SGD $\le 0.05$ (following~\cite{bian2019evaluation}). In the first three columns we report results obtained on exactly the same correspondences and number of iterations, making the outcomes directly comparable. For quick reference we also report the corresponding numbers from \cite{bian2019evaluation}.}
\vspace{-1em}
\begin{tabular}{c||c}
\label{table:SGD}
 & Obtained from~\cite{bian2019evaluation} \\
\begin{tabular}{l|c|c|c}
                 &  Our method (Sec.~\ref{sec:multiD}) & RANSAC & USAC-openCV  \\  
\hline
TUM              &  {62.8}  &   {60.9}     &  54.0   \\  
KITTI            &  {85.4}     &   {87.0}    & 83.8  \\  
T\&T             &   85.1    &    77.6       &    {88.1}      \\  
CPC              &  {50.3}  &   42.0     & 52.5    \\  
\end{tabular} &
\begin{tabular}{c|c|c}
  RANSAC &  USAC        & CF-RSC \\
\hline
      57.4             &    56.5        &  69.3    \\
       91.7            &    82.7     &  92.3    \\
       70.0            &    78.8     &  90.7     \\
        29.2           &  49.7          &  60.9      \\
\end{tabular} \\

\end{tabular}
\end{table}

\vspace{-3em}
\subsection{Comparisons against AQC-based quantum robust fitting}

Note that the AQC-based robust fitting methods of~\cite{doan2022hybrid,Farina_2023_CVPR} employed very different problem formulations to ours (see Sec.~\ref{sec:quantumsolutions}). A comparison with our GQC-based method would thus not be a meaningful comparison of quantum robust fitting. Moreover, the amenable problem sizes and solution quality of AQC are also limited~\cite{doan2022hybrid,Farina_2023_CVPR}, thus, conclusions drawn would not be scalable.

\section{Conclusions and discussion}\label{conclusion}

We demonstrated for the first time quantum robust fitting on a GQC using a novel quantum circuit for $\ell_\infty$ feasibility test in point fitting. We also showed how influence, a measure of outlyingness, can be accumulated to obtain influence for more complex estimation tasks. Results on a quantum simulator (Amazon Braket SV1) confirmed the algorithm’s correctness, while tests on a GQC (IonQ Aria) proved its feasibility on real quantum hardware. A fundamental matrix estimation benchmark indicated the promise of our method on a practical task.

Note that Alg.~\ref{alg:multiD} is a 
hybrid quantum-classical algorithm, similar to~\cite{doan2022hybrid,Farina_2023_CVPR}. While our method does not provide exponential speed-up to robust fitting, hybrid methods are nevertheless crucial to facilitate adoption of quantum computers in the near term~\cite{callison2022hybrid}. More broadly, developing quantum solutions for problems with existing classical solutions is a useful endeavor, since it can lead to new insights and techniques~\cite{shor2003haven}. In this spirit, our work represents an early but significant attempt at using GQC for a realistic computer vision problem, aiming to inspire more effective quantum methods for vision.

\vfill

\pagebreak

\bibliographystyle{splncs04}
\bibliography{egbib}
\pagebreak
\appendix

\section{Overview}
The supplementary materials provide implementation details on experiments mentioned in Sec. \ref{sec:qpuresults} in the main paper. Below is a brief overview of what is included in the document:
\begin{itemize}
  \item Sec. \ref{circuit}  outlines the specifics of the quantum circuits' design, including optimisation tricks.
  \item Sec. \ref{run} illustrates the way to execute the quantum circuits on both a quantum simulator and an actual quantum computer.
\end{itemize}

\section{Circuit design details}\label{circuit}
\begin{figure}
\centering
\begin{subfigure}{0.19\textwidth}
    \centering
    \begin{quantikz}[row sep={0.5cm,between origins}, column sep={0.3cm}]
        & \gate{H} & \\
    \end{quantikz}
    \captionsetup{width=2.5cm}
    \caption{Hadamard gate}
    \label{fig:h}
\end{subfigure}
\hfill
\begin{subfigure}{0.19\textwidth}
    \centering
    \begin{quantikz}[row sep={0.5cm,between origins}, column sep={0.3cm}]
        & \gate{X} & \\
    \end{quantikz}
    \captionsetup{width=2.5cm}
    \caption{Pauli-X gate}
    \label{fig:x}
\end{subfigure}
\hfill
\begin{subfigure}{0.19\textwidth}
    \centering
    \begin{quantikz}[row sep={0.5cm,between origins}, column sep={0.3cm}]
        & \ctrl{1} & \\
        & \targ{} & 
    \end{quantikz}
    \captionsetup{width=2.5cm}
    \caption{CNOT gate}
    \label{fig:cnot}
\end{subfigure}
\hfill
\begin{subfigure}{0.19\textwidth}
    \centering
    \begin{quantikz}[row sep={0.5cm,between origins}, column sep={0.3cm}]
        & \ctrl{2} & \\
        & \control{} & \\
        & \targ{} & 
    \end{quantikz}
    \captionsetup{width=2.5cm}
    \caption{CCNOT gate}
    \label{fig:ccnot}
\end{subfigure}
\begin{subfigure}{0.19\textwidth}
    \centering
    \begin{quantikz}[row sep={0.5cm,between origins}, column sep={0.3cm}]
        & \ctrl[wire style={"P(\pi/2)"}]{1} & \\
        & \control{} & \\
    \end{quantikz}
    \captionsetup{width=2cm}
    \caption{CPhase gate}
    \label{fig:cphase}
\end{subfigure}      
\caption{Quantum gates used in our circuits.}\vspace{-10pt}
\label{fig:gates}
\end{figure}
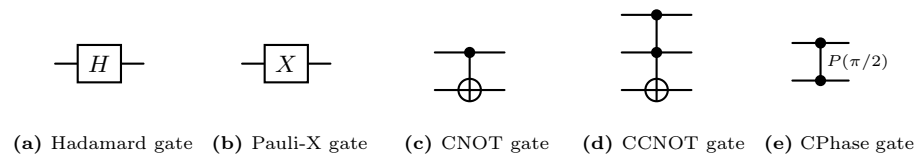

This section elaborates on the circuit design details and optimisation strategies not disclosed in the main paper, further extending the Circuit Design discussion in Sec. \ref{sec:approach} in the main paper. Please consult Fig. \ref{fig:gates} for representations of the quantum gates employed within our circuit design. The majority of gates utilized are detailed in  \cite[Chap.~5.2]{Nielsen_Chuang_2010}. All quantum circuits created for the experiments detailed in Sec. \ref{sec:qpuresults} adhere to the framework presented in Fig. \ref{fig:overview} of the main paper. Therefore, the following subsections will exclusively focus on the implementation of Block $D$, omitting the repetitive components from further illustration. The following subsections will illustrate 3 circuit instances. Sec. \ref{instance1} corresponds to the largest scale case discussed in Sec. \ref{sec:qpuresults} in the main paper, followed by Sec. \ref{instance2} and \ref{instance3}, where smaller instances and case-by-case optimisation are introduced.

\subsection{Circuit Instance 1}\label{instance1}
\begin{figure}[bt!]
\centering
\begin{subfigure}{0.8\textwidth}
    \includegraphics[width=\textwidth]{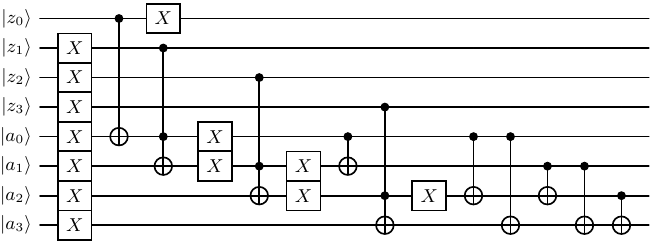}
    \caption{Block $A$}
    \label{fig:instance1_A}
\end{subfigure}
\vspace{9pt}
\begin{subfigure}{0.8\textwidth}
    \includegraphics[width=\textwidth]{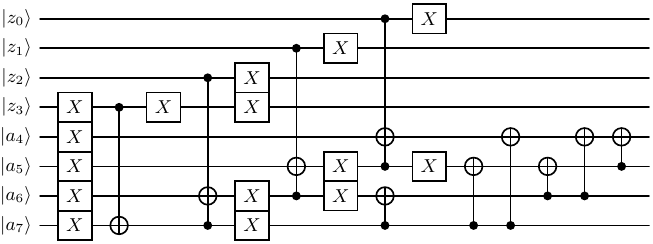}
    \caption{Block $B$}
    \label{fig:instance1_B}
\end{subfigure}
\vspace{9pt}
\begin{subfigure}{0.8\textwidth}
    \includegraphics[width=\textwidth]{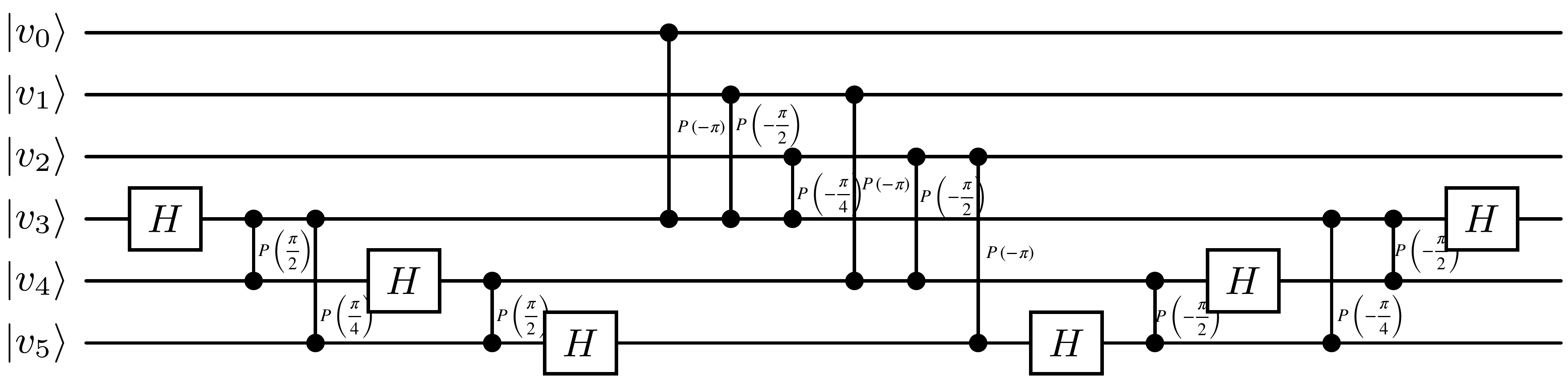}
    \caption{Block $S_1$}
    \label{fig:S1}
\end{subfigure}
\vspace{9pt}
\begin{subfigure}{0.4\textwidth}
    \includegraphics[width=\textwidth]{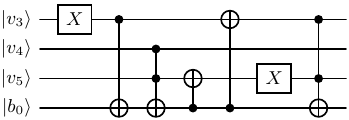}
    \caption{Block $S_2$}
    \label{fig:S2}
\end{subfigure}
\caption{Each block of Instance 1 is shown separately.}\vspace{-10pt}
\label{fig:instance1}
\end{figure}

This section corresponds to the largest experiment run in Sec. \ref{sec:qpuresults}, where input data $\cD = \{7,5,3,2\}$ at $C = 3$ bit precision and $2\epsilon = 2$. Due to the large scale of this particular circuit instance, the implementation of each block in Fig. \ref{fig:detailed} of the main paper is shown separately in Fig. \ref{fig:instance1}. Please note in this case Block $V_1$ and $V_2$ were illustrated in Fig. \ref{fig:v1v2} of the main paper. Instead of using QFT in $S_2$, a sequence of controlled-not gates can be used to minimise the circuit depth, at the expense of an additional auxiliary qubit $\ket{b_0}$ to carry the result of the feasibility test.

\subsection{Circuit Instance 2} \label{instance2}
\begin{figure}[h]\centering
    \includegraphics[width=0.7\columnwidth]{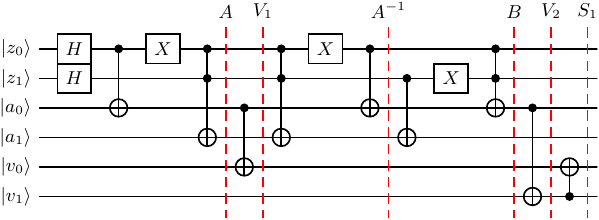}
    \caption{Instance 2}        
    \label{fig:instance2}
\end{figure}

This section corresponds to the case where input data $\cD = \{1, 0\}$ at $C = 1$ bit precision and $2\epsilon = 0$. See Fig. \ref{fig:instance2} for the implementation of Block $D$. In this case, the QFT part in Block $S_1$, and the whole block $S_2$ can be saved, as for $S_1$, the difference between the selected data can be computed from a CNOT gate, while for $S_2$, the resultant $\ket{{v_0}}$ from $S1$ can be directly passed as the feasibility test outcome.

\subsection{Circuit Instance 3}\label{instance3}
\begin{figure}
\centering
\begin{subfigure}{0.7\textwidth}
    \includegraphics[width=\textwidth]{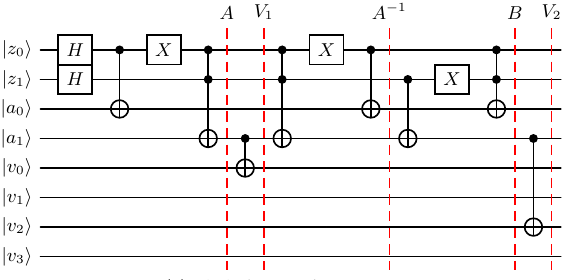}
    \caption{Block $A$ to $V_2$ of Instance 3.}
    \label{fig:instance3_A}
\end{subfigure}
\hfill
\begin{subfigure}{0.6\textwidth}
    \includegraphics[width=\textwidth]{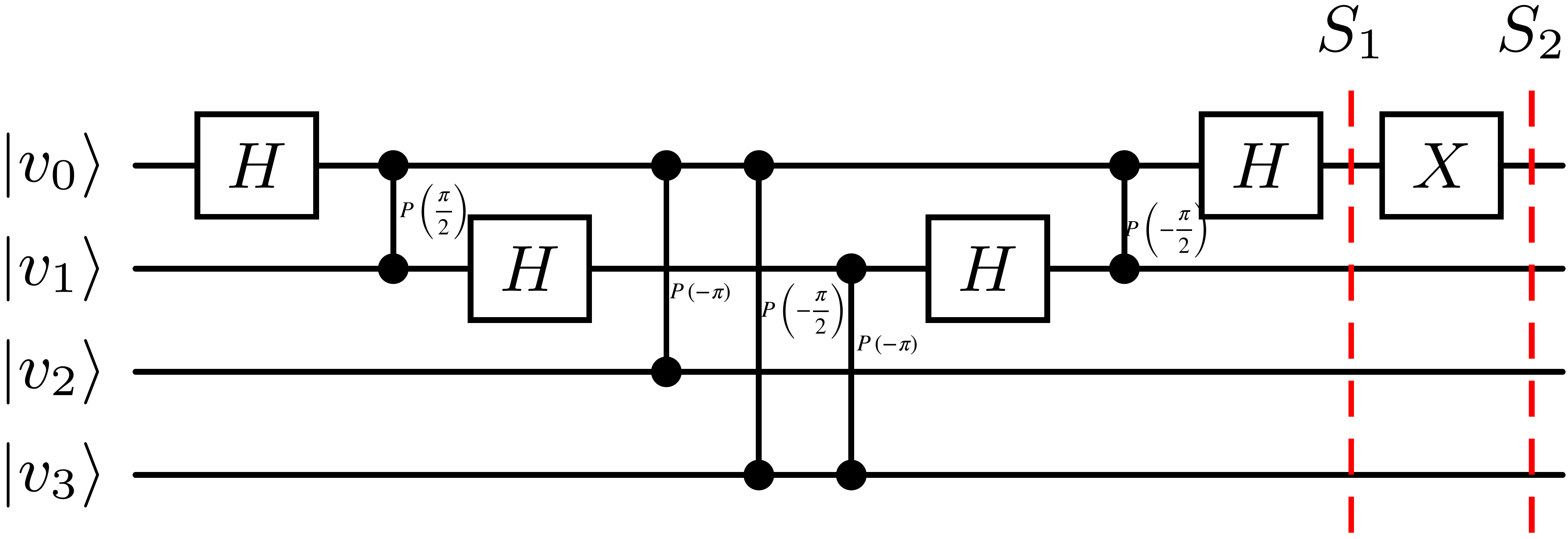}
    \caption{Block $S_1$ and $S_2$ of Instance 3.}
    \label{fig:instance3_B}
\end{subfigure}
\caption{Instance 3 split into 2 subplots.}
\label{fig:instance3}
\end{figure}

This section corresponds to the case where input data $\cD = \{2, 0\}$ at $C = 2$ bit precision and $2\epsilon = 1$. See Fig. \ref{fig:instance3} for the implementation of Block $D$. A simple Pauli-X gate is used in $S_2$ instead of QFT due to the specific computational scenario.

\subsection{Input \& output explained}
In accordance with Fig. \ref{fig:overview} and as detailed in Sec. \ref{sec:approach} of the main paper, the states $\ket{\bz}$, $\ket{\ba}$, and $\ket{\bv}$ are consistently initialized to $\ket{\mathbf{0}}$. The data input occurs through Blocks $V_1$ and $V_2$, as delineated in Fig. \ref{fig:v1v2}, where the states of $\ket{\bv}$ are selectively activated via CNOT gates. The focal point of our subsequent analysis is the output associated with $\ket{\bz}$. Upon completion, both $\ket{\ba}$ and $\ket{\bv}$ return to $\ket{\mathbf{0}}$, while $\ket{y}$ ends up to be $\ket{1}$. 

Considering that $\ket{\bz}$ achieves a uniform superposition subsequent to the application of the initial Hadamard gate, the resultant state of $\ket{\bv}$ encapsulates the influence sampled across all conceivable subsets of the input dataset. This naturally transits to the way we adopted for result analysis in Sec. \ref{sec:qpuresults}.

\section{Running quantum circuits} \label{run}
Experiments on our quantum circuits were conducted on both a quantum simulator (SV1) and a real quantum computer (IonQ Aria) via \path{qiskit_braket_provider} on Amazon Braket. \path{AWSBraketProvider} facilitates selection among various backends, encompassing both simulators and actual quantum computers. Please refer to code provided for details.

\end{document}